\pdfoutput=1

\documentclass[11pt]{article}

\usepackage[preprint]{acl}

\usepackage{times}
\usepackage{latexsym}

\usepackage[T1]{fontenc}

\usepackage[utf8]{inputenc}

\usepackage{microtype}

\usepackage{inconsolata}

\usepackage{amsmath}
\usepackage{amsfonts}
\usepackage{graphicx}
\usepackage{booktabs}
\usepackage{multirow}
\usepackage{threeparttable}
\usepackage{MnSymbol,bbding,pifont}
\usepackage{hyperref}

\title{NutePrune: Efficient Progressive Pruning with Numerous Teachers for Large Language Models}

\author{Shengrui Li\textsuperscript{1,2\textdagger}\ \ \ \ \ \ Junzhe Chen\textsuperscript{1,2\textdagger} \ \ \ \ \ \ Xueting Han\textsuperscript{2*} \ \ \ \ \ \ Jing Bai\textsuperscript{2}\\
  \textsuperscript{1}Tsinghua University\ \ \ \ \ \  \textsuperscript{2}Microsoft Research Asia \\
  \texttt{\{lsr22, chenjz20\}@mails.tsinghua.edu.cn} \\
  \texttt{\{chrihan, jbai\}@microsoft.com} \\
  }

\begin{document}
\maketitle

\begingroup\def\thefootnote{\textdagger}\footnotetext{Work done during internship at Microsoft Research Asia.}\endgroup
\begingroup\def\thefootnote{*}\footnotetext{Corresponding author.}\endgroup
\renewcommand{\thefootnote}{\arabic{footnote}}

\begin{abstract}
The considerable size of Large Language Models (LLMs) presents notable deployment challenges, particularly on resource-constrained hardware. Structured pruning, offers an effective means to compress LLMs, thereby reducing storage costs and enhancing inference speed for more efficient utilization. In this work, we study data-efficient and resource-efficient structure pruning methods to obtain smaller yet still powerful models.
Knowledge Distillation is well-suited for pruning, as the intact model can serve as an excellent teacher for pruned students. However, it becomes challenging in the context of LLMs due to memory constraints.
To address this, we propose an efficient progressive Numerous-teacher pruning method (NutePrune). NutePrune mitigates excessive memory costs by loading only one intact model and integrating it with various masks and LoRA modules, enabling it to seamlessly switch between teacher and student roles. This approach allows us to leverage numerous teachers with varying capacities to progressively guide the pruned model, enhancing overall performance. Extensive experiments across various tasks demonstrate the effectiveness of NutePrune. In LLaMA-7B zero-shot experiments, NutePrune retains 97.17\% of the performance of the original model at 20\% sparsity and 95.07\% at 25\% sparsity. Our code is available at \href{https://github.com/Lucius-lsr/NutePrune}{https://github.com/Lucius-lsr/NutePrune}.
\end{abstract}

\section{Introduction}

Large Language Models (LLMs) excel in language tasks \cite{openai2023gpt, touvron2023llama, thoppilan2022lamda, scao2022bloom}, but their substantial size poses deployment and inference challenges \cite{frantar2022gptq}. Techniques like model pruning \cite{molchanov2016pruning}, knowledge distillation \cite{ jiao2019tinybert}, and quantization \cite{dettmers2023qlora} have been proposed to address computational demands. The exploration of LLM pruning, especially structured pruning \cite{frantar2023sparsegpt}, holds great significance. Structured pruning reduces model size by removing coherent parameter groups, cutting inference costs on standard hardware. But it is more challenging than unstructured pruning in retaining the capabilities of LLMs \cite{hoefler2021sparsity}.
Existing methods either adopt data-efficient approaches, causing a performance decline \cite{ma2023llm}, or require extensive post-training to recover model performance \cite{xia2023sheared}. In this work, we investigate efficient methods to prune the model to higher sparsity without significant performance decline.

\begin{figure}
  \centering
  \includegraphics[width=0.5\textwidth]{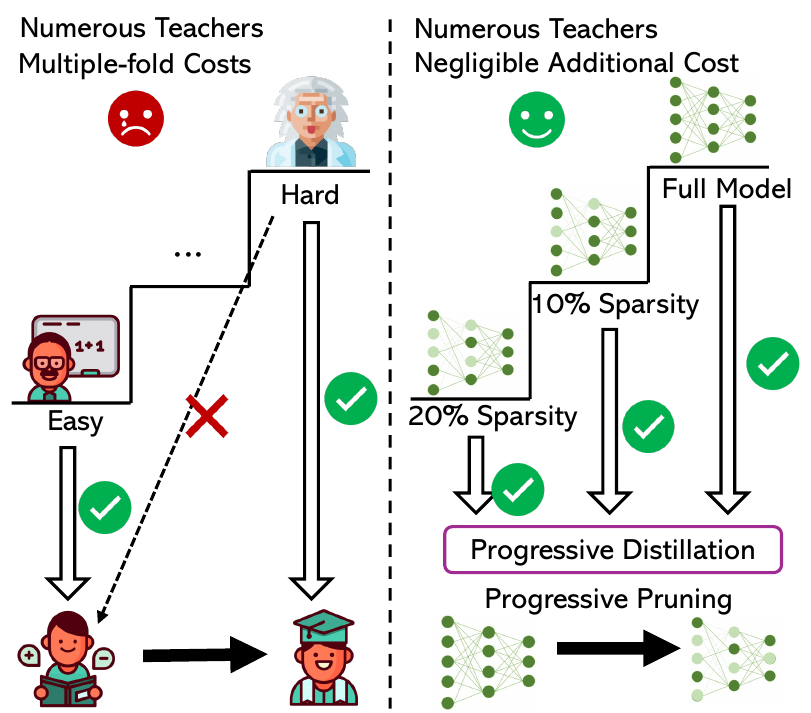}
  \caption{The advantage of our NutePrune. \textbf{Left}: Progressive distillation guides the student with teachers from easy to hard to avoid large capacity gap harming learning. But it suffers from multiple-fold costs of loading numerous teachers. \textbf{Right}: Our NutePrune leverages models with varying sparsity, enabling progressive distillation with negligible additional cost.}
  \label{fig_principle}
  \vspace{-0.5cm}
\end{figure}

Knowledge distillation (KD) aims to train a more compact student model with supervision from a larger teacher model \cite{sanh2019distilbert, gou2021knowledge}. 
It's widely adopted and proven highly effective in the field of LLMs. Progressive learning, utilizing intermediate teachers with a reduced gap in capabilities, has been demonstrated to improve performance in KD \cite{xiang2020learning}. 
Previous work has shown that pruning with a distillation objective can improve performance \cite{xia2022structured}. Distillation is particularly suitable for pruning since the full original model inherently serves as an excellent teacher for the pruned model \cite{sanh2020movement}, which can offer a more detailed supervisory signal than conventional supervised training, enhancing the effectiveness of pruning with limited data \cite{lagunas2021block}.

However, applying this method in the realm of LLMs proves challenging. Given the vastness of an LLM, loading it onto GPUs consumes a substantial amount of memory. Introducing an additional teacher model requires twice the memory, making it impractical with limited memory resources. Furthermore, relying on a single teacher may not be the best practice \cite{liu2020adaptive, wu2021one}. With the increasing gap of sparsity between teacher and student, the capacity gap is also widening, which toughens distillation. Employing multiple teachers with varying capacities can enhance the transfer of knowledge to students \cite{yuan2021reinforced}. However, when it comes to the distillation of LLMs, memory consumption of multiple teachers becomes an even more pressing concern.

\begin{table}[h!]
\small
  \centering
  \begin{tabular}{c|ccc}
    \toprule
    Method & NutePrune & LLM-Pruner & KD \\
    \midrule
    GPU Memory (GB) & 28.7 & 35.4 & 42.1 \\
    \bottomrule
    \end{tabular}
    \caption{\label{table-memory} GPU memory consumption during pruning.}
\end{table}

In this paper, we address the above challenges with an efficient progressive \textbf{Nu}merous-\textbf{te}acher pruning method (NutePrune). Our motivation is demonstrated in Figure \ref{fig_principle}. NutePrune aims to diminish the capacity gap between the full teacher model and the highly sparse student, thereby alleviating the difficulty of distillation \cite{su2021ernie, mukherjee2023orca, xiang2020learning}. Instead of relying solely on a single full teacher, we instruct the student with many teachers with varying sparsity. To achieve this, we formulate pruning as a optimization problem where we learn masks to prune sub-modules while updating model parameters through LoRA \cite{hu2021lora}. Specially, we load an intact model, serving dual roles as both a teacher and a student. In teacher mode, we incorporate the original model with collected frozen low-sparsity masks and corresponding LoRA modules. And in student mode, we incorporate it with learnable high-sparsity masks and LoRA modules. Since the masks and LoRA modules are highly parameter efficient, we collect and leverage numerous modules with different sparsity to incorporate numerous teachers and progressively prune the student. And as shown in Table \ref{table-memory}, this novel strategy remains highly memory efficient. Our contributions can be summarized as follows:

\begin{itemize}
    \item We propose a novel distillation method that progressively guide the student using numerous teachers with varying sparsity to narrow the capacity gap. Through progressive KD, we achieve higher model sparsity without significant performance decline on limited data.
    \item Our NutePrune only loads one intact model and switch it between teacher and student modes by incorporating various masks and LoRA modules. This novel efficient distilling method for pruning enables using numerous teachers and introduces no extra memory cost, which is especially critical for LLMs. 
    \item Extensive experiments, including LLaMA-1/2/3 with varying sizes and Mistral, demonstrate the effectiveness of our approach across perplexity, commonsense reasoning, MMLU, and BBH.
\end{itemize}

\section{Related Works}

\begin{table}[h!]
\small
  \centering
  \begin{tabular}{c|ccc}
    \toprule
    Pruning Type & Speedup & No Support & No Index \\
    \midrule
    Unstructured & & $\checkmark$ & \\
    Semi-Structured & $\checkmark$ & & \\
    Structured & $\checkmark \checkmark$ & $\checkmark$ & $\checkmark$ \\
    \bottomrule
    \end{tabular}
    \caption{\label{table-type} Structured pruning yield most significant speedup without any special hardware support or additional index storage.}
\end{table}
\vspace{-0.5cm}

\paragraph{Pruning for LLMs}
For LLMs, SparseGPT \cite{frantar2023sparsegpt} and WANDA \cite{sun2023simple} employ unstructured pruning methods, while N:M sparsity \cite{zhou2021learning} is considered semi-structured. Despite the effectiveness of these methods, their intricate structures do not yield significant inference speedup on standard hardware \cite{frantar2023sparsegpt} and they need to store additional indexes. As compared in Table \ref{table-type}, structured pruning offers significant advantages, resulting in increased focus on this field in recent works. CoFi \cite{xia2022structured} and nn pruning \cite{lagunas2021block} are proposed for smaller language models like BERT \cite{devlin2018bert}, often designed for specific tasks. CoFi loads both the teacher and student models, which is impractical for LLMs. Sheared-LLaMA \cite{xia2023sheared} proposes pruning LLMs using a dynamic pre-training method, enhancing performance through extensive data and training resources. 

However, concerns persist regarding limited memory and training resources for LLMs. In a pioneering effort, LLM-Pruner \cite{ma2023llm} prunes LLMs in one-shot and utilizes LoRA \cite{hu2021lora} for fine-tuning. LoRAPrune \cite{zhang2023pruning} employs iterative pruning, replacing gradients on full weights with gradients on LoRA to calculate group importance. Compresso \cite{guo2023compresso} leverages LoRA and elaborately designed prompts for training and inference. Meanwhile, LoRAShear \cite{chen2023lorashear} employs LoRA and a dynamic fine-tuning scheme to recover knowledge.

\paragraph{Knowledge Distillation (KD) for LLMs}
KD \cite{hinton2015distilling} has emerged as a vital technique to reduce inference costs while maintaining performance quality in the context of LLMs. Prior work of KD \cite{taori2023stanford, fu2023specializing} mostly focuse on black-box KD, using teacher's generations to fine-tune the student. With the rise of open-source LLMs \cite{zhang2022opt, touvron2023llama}, interest in white-box KD is growing. White-box KD, leveraging teacher weights and logits, provides richer supervision signals, enhancing language abilities \cite{agarwal2023gkd, gu2023knowledge, wen2023f}. Despite progress on small language models, significant performance gaps between large and small models persist \cite{achiam2023gpt, anil2023palm}.

Progressive knowledge distillation \cite{xiang2020learning} has proven effective by using intermediate teachers to bridge the capacity gap with LLMs, especially in scenarios reliant on data generated by multiple teachers \cite{mukherjee2023orca}. Orca \cite{mukherjee2023orca} first learns from easier examples from ChatGPT and then from harder ones from GPT-4, enhancing performance for smaller students in KD. However, applying white-box KD to LLMs poses challenges due to substantial memory requirements for loading both teacher and student models. This challenge becomes even more difficult when attempting to load multiple teachers.


\section{Methodology}

\begin{figure*}[h!]
  \centering
  \includegraphics[width=\textwidth]{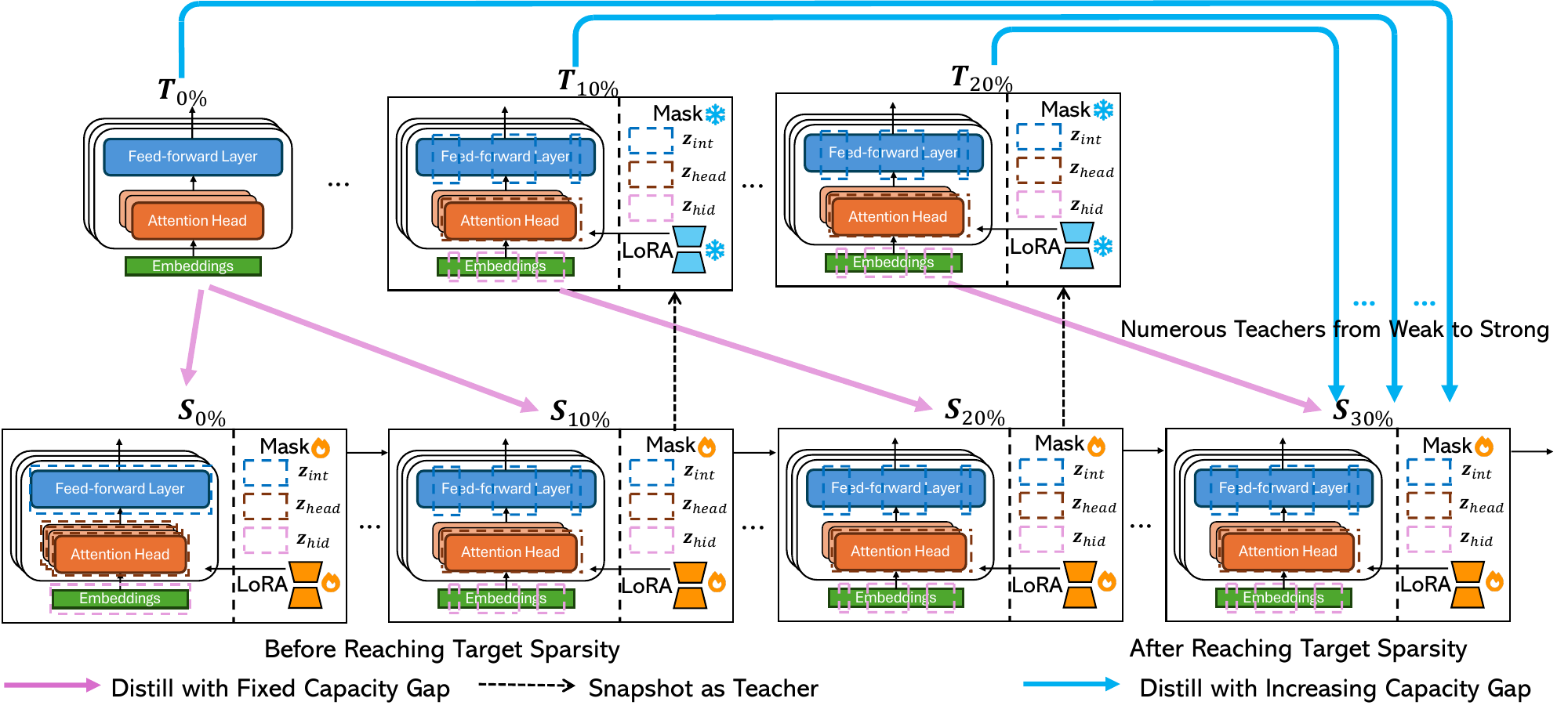}
    \vspace{-0.5cm}
  \caption{The overall framework of NutePrune. The pruned model is frozen and incorporated with learnable masks and LoRA. During pruning, the model is guided by numerous teachers. Before pruned to the target sparsity (e.g. 30\%), it learns from teachers with a fixed capacity gap. Once the target sparsity is achieved, it continues to learn from all previous teachers from weak to strong. All these teachers are derived from snapshots of the student model itself. Since only the mask and LoRA modules are snapshotted, the additional memory cost is negligible.}
  \label{fig_framework}
\vspace{-0.5cm}
\end{figure*}

In this section, we first introduce how our NutePrune enables efficient knowledge distillation for structured pruning in \ref{method_kd}. Then, to narrow capacity gap during distillation, we introduce the progressive knowledge distillation method that collects and incorporates numerous teachers in \ref{method_progressive}. The overview framework is illustrated in Figure \ref{fig_framework}.

\subsection{Efficient Distillation for Structured Pruning}
\label{method_kd}

We formulate structure pruning as a constrained optimization problem where we simultaneously learn masks to prune the structure and update the model to recover ability. To mitigate memory consumption, we utilize LoRA for model updates, making pruning the process of training these masks and LoRA parameters.  

\paragraph{Learning masks to control the pruned structure}

Three types of structure are pruned: attention heads, FFN intermediate dimensions, and hidden dimensions. We achieve this by learning masks $\mathbf{z}_{head}, \mathbf{z}_{int}, \mathbf{z}_{hid} \in \{0,1\}$. Formally, the multi-head attention module $\operatorname{MHA}(x)$ and feed-forward networks $\operatorname{FFN}(x)$ of layer $l$ are pruned as:

\begin{equation}
    \operatorname{MHA}^l(X) = \mathbf{z}_{hid}\cdot \sum_{h=1}^{N_{head}} \mathbf{z}_{head}^{l,h} \operatorname{Att}^{l,h}(X).
\end{equation}

\begin{equation}
    \operatorname{FFN}^l(X) = \mathbf{z}_{hid}\cdot W_D^l\left(\mathbf{z}_{int}^l\cdot W_U^l(X)\cdot W_G^l(X)\right)
\end{equation}
where $\operatorname{Att}()$ is the attention module and activation is omitted. $W_D, W_U, W_G$ are down projection, up projection, and gating projection.

During mask training, we calculate the remaining size to obtain the expected sparsity $\hat{s}$:
\begin{equation}
\begin{split}
    \hat{s}(\mathbf{z}) &= \frac{1}{M}\cdot 4 \cdot d_h \cdot \sum_l^L \sum_h^{N_{head}} \sum_k^d \mathbf{z}_{head}^{l,h} \mathbf{z}_{hid}^k\\
    &+\frac{1}{M}\cdot 3 \cdot \sum_l^L \sum_i^{d_{int}} \sum_k^d \mathbf{z}_{int}^{l,i} \mathbf{z}_{hid}^k,
\end{split}
\end{equation}
where $M$ denotes full model size. $L$ is number of layers. $d_h, N_{head}, d, d_{int}$ are head dimension, number of head, hidden dimension, and intermediate dimension, correspondingly.

All masking variables are learned as real numbers in $[0, 1]$ during training. We follow \cite{louizos2017learning, guo2023compresso} and employ the augmented $L_0$ regularization, which is detailed in Appendix \ref{sec:appendix_l0}.

\paragraph{Updating parameters with LoRA}

Considering massive memory usage during full fine-tuning for LLMs, we incorporate lightweight LoRA \cite{hu2021lora} modules into LLM weights to update parameters during pruning.

An incorporated module $W'$ is consisted of the original weight $W:\mathbb{R}^{n} \rightarrow \mathbb{R}^{m}$ and sequential LoRA weights parallel to $W$:
\begin{equation}
    W'(X) = W(X) + W_B\left(W_A(X)\right),
\end{equation}
where $W_A:\mathbb{R}^{n} \rightarrow \mathbb{R}^{r}, W_B:\mathbb{R}^{r} \rightarrow \mathbb{R}^{m}$ and $r \ll m, n$. During training, $W$ is frozen and only $W_A$ and $W_B$ are learnable.

\paragraph{Efficient distillation}
Instead of simultaneously loading two massive models into memory, we propose to incorporate the frozen and intact model $M$ with different lightweight masks and LoRA modules for the teacher and the student. Formally, let $\mathbf{I}=\{\mathbf{z}, \mathbf{W}_A,  \mathbf{W}_B\}$ denotes the set of all masks and LoRA modules which is highly parameter efficient ($|\mathbf{I}|\ll |\mathbf{M}|$). By incorporating $\mathbf{I}$ into $\mathbf{M}$, we obtain $\mathbf{M}_\mathbf{I}$. The objective of knowledge distillation is the KL-divergence \cite{van2014renyi} between teacher's and student's output probability distributions $p$:
\begin{equation}
    \mathcal{L}_{KL}=D_{KL}(p(\mathbf{M}_{\mathbf{I}_S}, x),p(\mathbf{M}_{\mathbf{I}_T}, x)),
\end{equation}
where $x$ denotes training data. $\mathbf{I}_S$ and $\mathbf{I}_T$ denote the lightweight modules of student and teacher.

Additionally, intermediate layers of a teacher model can serve as effective targets for training a student model \cite{chen2021cross}. This objective can be formulated as:
\begin{equation}
    \mathcal{L}_{layer}=\sum_l^L \operatorname{MSE}(\mathbf{h}_l(\mathbf{M}_{\mathbf{I}_S}, x), \mathbf{h}_l(\mathbf{M}_{\mathbf{I}_T}, x)),
\end{equation}
where $\mathbf{h}_l$ is the hidden embedding of the $l$-th layer. Therefore, the overall objective is:
\begin{equation}
    \mathcal{L} = \mathcal{L}_{KL} + \alpha_1 \mathcal{L}_{layer} + \alpha_2 \mathcal{L}_0,
\end{equation}
where $\alpha_1, \alpha_2$ are hyperparameters to control the importance of different loss terms.

\subsection{Progressive Knowledge Distillation with Numerous Teachers}
\label{method_progressive}

\begin{figure}
  \centering
  \includegraphics[width=0.45\textwidth]{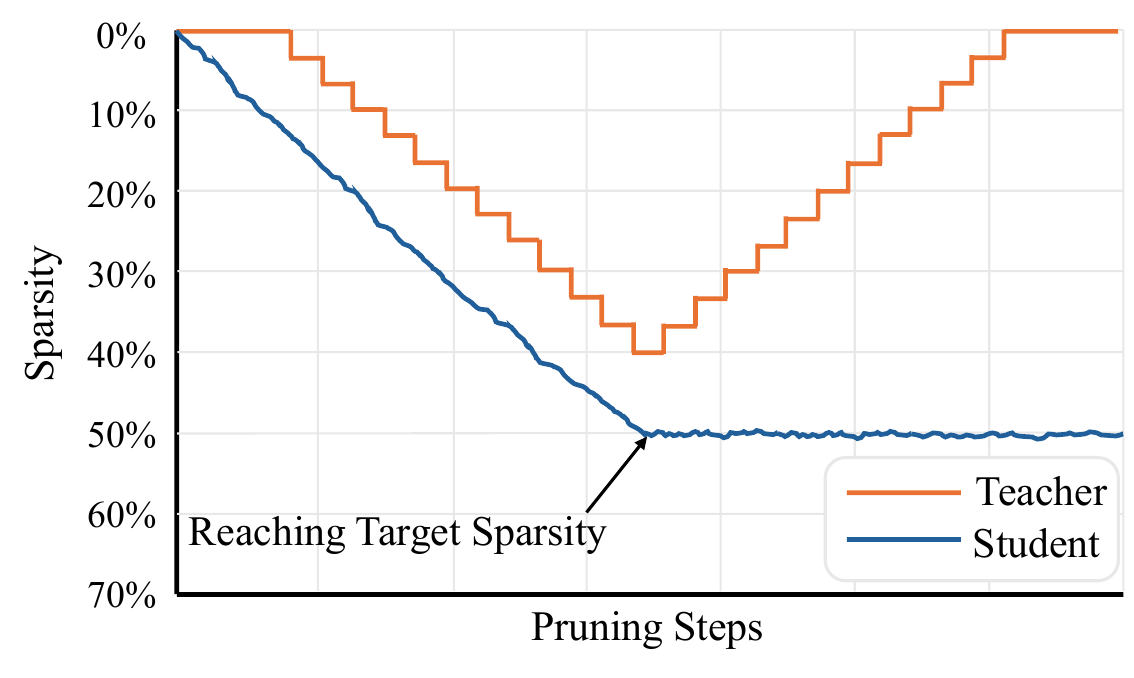}
  \caption{Illustration of the sparsity of teacher and student models during pruning. Take the example with the target sparsity $t=50\%$ and sparsity gap $g=10\%$.}
  \label{fig_sparsity}
\end{figure}

All teachers are collected from the snapshot of students as the dotted line illustrated in Figure \ref{fig_framework}. To narrow the capacity gap between the intact teacher and high sparsity students, we leverage a novel progressive knowledge distillation (PKD) method for pruning. It consists of two stages when pruning a model from 0\% sparsity as illustrated in Figure \ref{fig_sparsity}.

\paragraph{Before reaching target sparsity}
The sparsity of pruned model gradually increase from 0 to $t$. To narrow the sparsity gap, we set a fixed gap value $g$ and make the pruned model $S$ guided by teachers $T$ whose sparsity $\hat{s}(T)$ is approximately $g$ less than $\hat{s}(S)$: $\hat{s}(T)=\hat{s}(S)-g$. These teachers are snapshots of previous students. The original intact model serves as the teacher for student $\hat{s}(S)<g$.

To avoid collecting too many teachers, we only collect teachers with an interval of $i$. Therefore, for any teacher with sparsity $\hat{s}(T)$, it is responsible for guiding a student set within a range of sparsity. We use $\rightarrow$ to denote the relationship in which a teacher distills knowledge to students.
\begin{equation}
    T \rightarrow \{S|\hat{s}(T)+g<\hat{s}(S)<\hat{s}(T)+g+i\}.
\end{equation}
And the intact model $\mathbf{M}=T_0$ is responsible for the early students whose sparsity is less than $g+i$:
\begin{equation}
    T_0 \rightarrow \{S|\hat{s}(S)<g+i\}.
\end{equation}

\paragraph{After reaching target sparsity}
When the pruned model reaches the target sparsity $t$, we proceed to the second stage of PKD. The model undergoes distillation by all preceding teachers, with a reduction of sparsity in the teachers. This gradual process guides the model's learning trajectory from weaker to stronger knowledge and from easier to more challenging concepts. Throughout this stage, the sparsity of the pruned model $\hat{s}$ remains close to the target sparsity $t$, while the masks $\mathbf{z}$ and LoRA modeuls $\mathbf{W}_A,  \mathbf{W}_B$ are continually optimized.

To receive sufficient instruction from the best model (the intact model $\mathbf{M}$), the teacher model is maintained as $\mathbf{M}$ during the final period.

\subsection{Post Fine-tuning}
After the pruning phase, to obtain better performance, we undergo a post fine-tuning stage following LLM-Pruner \cite{ma2023llm}. We fix the masks and only fine-tune LoRA modules on the Standford Alpaca \cite{taori2023stanford} dataset.

\section{Experiments}

\subsection{Experimental Setup}

\paragraph{Datasets}
To assess the zero-shot ability of LLMs, we perform zero-shot classification tasks on seven commonsense reasoning benchmarks: BoolQ \cite{clark2019boolq}, PIQA \cite{bisk2020piqa}, HellaSwag \cite{zellers2019hellaswag}, WinoGrande \cite{sakaguchi2021winogrande}, ARC-easy \cite{clark2018think}, ARC-challenge \cite{clark2018think}, and OpenBookQA (OBQA) \cite{mihaylov2018can}.
We evaluate the general capcability of LLMs on the perplexity metric with WikiText \cite{merity2016pointer}.
Additionally, to evaluate the in-context learning ability, We report the results on 5-shot MMLU \cite{hendrycks2020measuring}, and 3-shot BBH \cite{suzgun2022challenging}. 

\paragraph{Models}
NutePrune is applicable across various models of different sizes. We assess the performance of NutePrune on the LLaMA-1 family (7B/13B) \cite{thoppilan2022lamda}, LLaMA-2 \cite{touvron2023llama} family (7B/13B), LLaMA-3-8B and Mistral-7B \cite{jiang2023mistral}. 


\paragraph{Baselines}
Considering the benefits of inference acceleration, we focus on structured pruning. We first replicate conventional methods: Magnitude pruning (MaP) \cite{li2018optimization}, Movement Pruning (MvP) \cite{sanh2020movement}, and WANDA \cite{sun2023simple}. For recent open-source methods, we implement LLM-Pruner \cite{ma2023llm} and Compresso \cite{guo2023compresso} and conduct detailed comparisons. For more recent works that are not publicly available, we assess NutePrune using the same settings as theirs. This includes LoRAPrune \cite{zhang2023pruning} and LoRAShear \cite{chen2023lorashear}.

\paragraph{Implementation details} For pruning stage, we sample 20,000 sentences from the C4 \cite{raffel2020exploring} dataset with a length of 512 tokens. We train with AdamW optimizer, a batch size of 16, and learning rates of 0.1 for masks and 0.001 for LoRA. We prune the model for 7 epochs 
and a linear sparsity schedule for target sparsity warmup: 4 epochs for 20\% sparsity and 1 epoch for 50\%. 
The sparsity gap between the teacher and student $g$ is 10\% and the snapshot interval $i$ of teachers is 1\%. After pruning, we post fine-tune the pruned model on the Alpaca dataset \cite{taori2023stanford} for 3 epochs.
All experiments are conducted on one A100 GPU (80G).

\begin{table*}[h!]
  \centering
  \small
  \begin{threeparttable}
  \begin{tabular}{cc|c|ccp{0.33in}p{0.33in}p{0.35in}p{0.35in}p{0.33in}|cc}
    \toprule
    Ratio & Method & \scriptsize{WikiText2$\downarrow$} & BoolQ & PIQA & \scriptsize{HellaSwag} & \scriptsize{WinoGrande} & ARC-e & ARC-c & OBQA & Avg. & $\star$Avg.\\
    \midrule
    0\% & LLaMA-7B & 5.68 &  73.18 & 78.35 & 72.99 & 67.01 & 67.45 & 41.38 & 42.40 & 63.25 & 66.39 \\
        \hline
    \multirow{3}{*}{20\%} & LLM-Pruner & 9.96 &  59.39 & 75.57 & 65.34 & 61.33 & 59.18 & 37.12 & 39.80 & 56.82 & 59.01 \\
    ~  & LoRAPrune & - & 57.98 & 75.11 & 65.81 & 59.90 & 62.14 & 34.59 & 39.98 & 56.50 & - \\
    ~  & $\dagger$\textbf{NutePrune} & \textbf{8.02} & 63.21 & 76.55 &	67.96 &	66.69 &	63.72 &	38.05 &	40.00 &	\textbf{59.46} & \textbf{63.03}\\
    \hline
    \multirow{9}{*}{\shortstack{20\%\\Tuned}} & MaP & 12.67& 60.00	& 76.12 & 	65.43 &	60.93 &	60.31 &	37.80 &	39.80 & 57.20 & 60.05\\
    ~ & MvP & 10.52 & 64.50	& 73.50 &	62.50 &	61.80 &	62.42 &	36.95 &	37.80 &	57.07 & 58.76\\
    ~ & WANDA & - & 65.75 & 74.70 & 64.52 & 59.35 & 60.65 & 36.26 & 39.40 & 57.23 & -\\
    ~  & LLM-Pruner & 8.57 & 69.54 & 76.44 & 68.11 & 65.11 & 63.43 & 37.88 & 40.00 & 60.07 & 61.94 \\
    ~  & LoRAPrune & - & 65.82 & 79.31 & 70.00 & 62.76 & 65.87 & 37.69 & 39.14 & 60.05 & - \\
    ~  & LoRAShear & - & 70.17 & 76.89 & 68.69 & 65.83 & 64.11 & 38.77 & 39.97 & 60.63 & - \\
    ~  & Compresso & 10.38 & 73.64 & 75.08	& 64.77	& 67.72	& 66.12	& 37.54	& 40.40	& 60.75 & 62.60 \\
    ~  & $\ddagger$\textbf{NutePrune} & 8.04 & 72.69 & 76.71 & 68.99	& 65.51 & 65.49	& 38.48	& 40.20	& 61.15 & 63.57 \\
    ~  & \textbf{NutePrune} & \textbf{7.65} & 74.56 & 77.04 & 70.01 & 65.67 & 65.78 & 37.97 & 39.20 & \textbf{61.46} & \textbf{64.39} \\
    \hline
    25\% & $\dagger$\textbf{NutePrune} & 9.04 & 68.10 & 75.35 & 66.75 & 62.04 & 58.08 & 36.77 & 39.00 & 58.01 & 61.72 \\
    \hline
    \multirow{2}{*}{{\shortstack{25\%\\Tuned}}} & $\ddagger$\textbf{NutePrune} & - & 65.84 & 76.17 & 66.69 & 64.56 & 61.49 & 37.03 & 39.20 & 58.71 & 63.12 \\
    ~ & \textbf{NutePrune} & 7.85  & 68.99 & 77.20 & 67.90 & 65.04 & 63.76 & 37.80 & 40.20 & 60.13 & 63.78 \\
        \hline
    \multirow{3}{*}{50\%} & LLM-Pruner& 98.10  & 52.32& 59.63 & 35.64 & 53.20 & 33.50 & 27.22 & 33.40 & 42.13 & 40.94 \\
    ~  & LoRAPrune & - & 51.78 & 56.90 & 36.76 & 53.80 & 33.82 & 26.93 & 33.10 & 41.87 & - \\
    ~ & $\dagger$\textbf{NutePrune} & \textbf{17.45} & 62.29	& 67.95	& 53.03	& 57.06	& 45.45	& 30.03	& 36.60	& \textbf{50.35} & \textbf{53.14} \\
    \hline
    \multirow{9}{*}{\shortstack{50\%\\Tuned}} & MaP & 33.18 & 39.69 & 66.81 &	42.49 &	50.67 &	49.32 &	30.63 &	31.40 &	44.43 & 46.33\\
    ~ & MvP & 27.62 & 59.94 & 	63.06 &	40.98 &	55.64 &	44.07 &	26.79 &	31.80 &	46.04 & 46.23\\
    ~ & WANDA & - & 50.90 & 57.38 & 38.12 & 55.98 & 42.68 & 34.20 & 38.78 & 45.43 & - \\
    ~  & LLM-Pruner & 22.76  & 61.47 & 68.82 & 47.56 & 55.09 & 46.46 & 28.24 & 35.20 & 48.98  & 48.97 \\
    ~  & LoRAPrune & -  & 61.88 & 71.53 & 47.86 & 55.01 & 45.13 & 31.62 & 34.98 & 49.71 & -\\
    ~  &LoRAShear & - & 62.12 & 71.80 & 48.01 & 56.29 & 47.68 & 32.26 & 34.61 & 50.39 & - \\
    ~  &Compresso & 59.73 & 60.09 & 66.70 & 39.31 & 51.93 & 48.82 & 27.82 & 33.40 & 46.87 & 47.43 \\
    ~ & $\ddagger$\textbf{NutePrune} & 16.72 & 62.20 & 69.91 & 53.87 & 57.77 & 46.59 & 31.74 & 35.80 & 51.13 & 53.94 \\
    ~ & \textbf{NutePrune} & \textbf{13.20} & 62.26 & 71.00 & 55.88 & 57.54 & 51.68 & 32.17 & 34.40 & \textbf{52.13} & \textbf{54.91} \\
    \bottomrule
    \end{tabular}
        \begin{tablenotes}
    \item $\dagger$ only prunes the model by training masks without incorporating LoRA modules.
    \item $\ddagger$ prunes the model by co-training the masks and LoRA modules but without post fine-tuning on Alpaca.
    \item $\star$ includes results with the newer version of \textit{lm-evaluation-harness}. See Appendix \ref{sec:appendix_new} for detail.
    \end{tablenotes}
    \end{threeparttable}
    \caption{\label{table-main}Performance (\%) of the compressed LLaMA-7B models.}
\end{table*}

\begin{table*}[h!]
  \centering
  \small
  \begin{threeparttable}
  \begin{tabular}{cc|c|ccp{0.4in}p{0.5in}p{0.35in}p{0.35in}p{0.33in}|cc}
    \toprule
    Ratio  & Method & WikiText2$\downarrow$ & BoolQ & PIQA & HellaSwag & WinoGrande & ARC-e & ARC-c & OBQA & $\star$Avg.\\
    \midrule
    0\% & LLaMA-2-7B & 5.47 & 77.74	& 79.11	& 75.97	& 69.06	& 76.35 &	46.33 &	44.20 &	66.97 \\
        \hline
    \multirow{2}{*}{20\%} &  LLM-Pruner & 12.94 & 50.55 &	75.46 &	67.18 &	65.67 &	67.38 &	38.14 &	38.40 & 57.54 \\
    ~  & \textbf{NutePrune} & \textbf{8.74} & 77.06 &	76.66 &	70.56 &	65.59 &	71.97 &	42.58 &	42.40 &	\textbf{63.83} \\
    \hline
    \multirow{2}{*}{50\%} &  LLM-Pruner & 24.47 & 54.13	& 68.06	& 46.71	& 51.54	& 50.97	& 25.85	& 34.00	& 47.30 \\
    ~  & \textbf{NutePrune} & \textbf{12.94} & 66.24	& 70.83	& 57.04	& 59.51	& 58.46	& 31.97	& 34.00	& \textbf{54.01} \\
    \bottomrule
    \end{tabular}
    \end{threeparttable}
    \caption{\label{table-llama2}Performance (\%) of the compressed LLaMA-2-7B models.}
\end{table*}

\begin{table*}[h!]
  \centering
  \small
  \begin{threeparttable}
  \begin{tabular}{cc|c|ccp{0.4in}p{0.5in}p{0.35in}p{0.35in}p{0.33in}|cc}
    \toprule
    Ratio  & Method & WikiText2$\downarrow$ & BoolQ & PIQA & HellaSwag & WinoGrande & ARC-e & ARC-c & OBQA & $\star$Avg.\\
    \midrule
    0\% & Mistral-7B &  5.25 & 83.73 &	82.26 &	81.05 &	74.19 &	80.89 &	53.84 &	43.80 &	71.39\\
    \hline
    \multirow{2}{*}{20\%} &  LLM-Pruner & 7.50 &  77.52	& 78.13 &	73.64 &	69.46 &	72.59 &	46.41 &	41.80 &	65.65\\
    ~  & \textbf{NutePrune} & \textbf{7.06} & 78.29	& 80.41 &	75.57 &	68.82 &	76.35	& 45.05 & 42.20 &	\textbf{66.67}  \\
    \hline
    \multirow{2}{*}{50\%} &  LLM-Pruner & 30.51 & 62.48 &	66.59 &	48.00 &	56.51 &	52.61 &	28.07 &	29.80 &	49.15  \\
    ~  & \textbf{NutePrune} & \textbf{12.29} & 63.64 &	72.63 &	57.95 &	61.25 &	62.46 &	35.41 &	33.80 &	\textbf{55.31} \\
    \bottomrule
    \end{tabular}
    \end{threeparttable}
    \caption{\label{table-mistral}Performance (\%) of the compressed Mistral-7B models.}
\end{table*}

\subsection{Results}

\paragraph{Zero-shot performance}

\begin{table*}[h!]
  \centering
  \small
  \begin{threeparttable}
  \begin{tabular}{cc|c|ccp{0.4in}p{0.5in}p{0.35in}p{0.35in}p{0.33in}|cc}
    \toprule
    Ratio  & Method & WikiText2$\downarrow$ & BoolQ & PIQA & HellaSwag & WinoGrande & ARC-e & ARC-c & OBQA & $\star$Avg.\\
    \midrule
    0\% & LLaMA-3-8B & 6.14 & 81.04 &	80.85 &	79.18 &	73.40 &	80.13 &	53.16 &	44.60 &	70.34 \\
    \hline
    \multirow{2}{*}{20\%} &  LLM-Pruner & 10.06 & 71.50 &	77.97 &	70.49 &	68.75 &	72.35 &	42.83 &	38.40 &	63.18 \\
    ~  & \textbf{NutePrune} & \textbf{9.51} & 78.65 &	79.76 &	73.74 &	70.09 &	76.22 &	45.90 &	43.40 &	\textbf{66.82} \\
    \hline
    \multirow{2}{*}{50\%} &  LLM-Pruner & 27.37 & 41.59 &	67.46 &	46.53 &	55.64 &	49.71 &	27.22 &	31.60 & 45.68  \\
    ~  & \textbf{NutePrune} & \textbf{21.72} & 65.20 &	68.72 &	52.30 &	58.64 &	53.41 &	29.44 &	35.00 &	\textbf{51.82} \\
    \bottomrule
    \end{tabular}
    \end{threeparttable}
    \caption{\label{table-llama3}Performance (\%) of the compressed LLaMA-3-8B models.}
\end{table*}

\begin{table*}[h!]
  \centering
  \small
  \begin{tabular}{cc|cccc}
    \toprule
    Ratio & Method & Avg. Zero-Shot (\%) & WikiText2$\downarrow$ & MMLU (5-shot) & BBH (3-shot) \\ 
    \midrule
    0\% & LLaMA-13B & 67.63 & 5.62 & 46.90 & 37.72 \\ 
    \hline
    \multirow{2}{*}{20\%} & LLM-Pruner & 65.76 & 6.95 & 29.57 & 15.77\\ 
    ~ & NutePrune & \textbf{67.51} & \textbf{6.55} & \textbf{39.37} & \textbf{31.99}\\ 
        \hline
        \hline
    0\% & LLaMA-2-13B & 68.80 & 5.30 & 54.86 & 39.53\\ 
    \hline
    \multirow{2}{*}{20\%} & LLM-Pruner & 65.87 & 7.25 & 31.26 & 25.20\\ 
    ~ & NutePrune & \textbf{67.39} & \textbf{6.86} & \textbf{45.78} & \textbf{31.22}\\ 
    \bottomrule
    \end{tabular}
    \caption{\label{table-13b}Performance of the compressed LLaMA-13B and LLaMA-2-13B models with 20\% sparsity.}
\end{table*}

Table \ref{table-main} demonstrates PPL and zero-shot performances on commonsense reasoning tasks for compressed LLaMA-7B models. 
The reported results include experiments for 20\%, 25\% and 50\% sparsity levels, covering scenarios with and without parameter tuning.

The average performance of NutePrune consistently outperforms previous methods across all settings. 
For pruning without tuning, NutePrune outperforms LLM-Pruner by 2.64\%/8.22\% at 20\%/50\% sparsity, underscoring its ability to derive a more effective pruned structure compared to other methods.
For pruning with LoRA con-trained, NutePrune improves from 59.46\%/50.35\% to 61.46\%/52.13\% at 20\%/50\% sparsity, indicating co-training with LoRA could help recover model capability damaged by pruning.
And with additional post fine-tuning on Alpaca, notably, it retains 97.17\% of the performance of the original model at 20\% sparsity and 95.07\% at 25\% sparsity.

Table \ref{table-llama2}, \ref{table-mistral} and \ref{table-llama3} further demonstrates performances for compressed LLaMA-2-7B, Mistral-7B and LLaMA-3 models. At the same sparsity, multi-query attention models experience a more significant performance decline. Nevertheless, NutePrune consistently outperforms LLM-Pruner, proving our method is applicable across various models. Noticeable improvements are observed at higher sparsity levels, proving the effectiveness of our PKD in mitigating the capacity gap during distillation.

\paragraph{Pruning of larger model}
We assess larger models: LLaMA-13B and LLaMA-2-13B with 20\% sparsity. To evaluate the ability of these stronger models, we further assess their in-context learning ability with MMLU and BBH \cite{brown2020language}. As demonstrated in Table \ref{table-13b}, our approach yield an average zero-shot commonsense reasoning performance of 67.51\% and 67.39\%, which is only slightly lower than the full model and much higher than LLM-Pruner. It also outperforms LLM-Pruner in terms of PPL in WikiText2. For in-context learning ability, NutePrune achieves a score of 36.37 MMLU and 31.99 BBH in LLaMA-13B, and 45.78 MMLU and 31.22 BBH in LLaMA-2-13B. The slight decline in performance compared to the full model is acceptable, indicating that NutePrune maintains sufficient in-context learning capability. Additionally, when compared to LLM-Pruner, our advantages are clearly evident.

\paragraph{Inference latency}
We test the inference latency by generating from 64 tokens to 256 tokens on vLLM \cite{kwon2023efficient}, which is a fast and widely deployed library for LLM inference and serving. The results are presented in Table \ref{table-speed}. NutePrune achieves latency savings of 11\% and 29\% at sparsity levels of 20\% and 50\%. While LLM-Pruner save slightly more latency due to its predefined neater structure, it comes at the cost of reduced flexibility in tailoring. As sparsity increases, the difference becomes negligible.
\begin{table}[h!]
  \centering
  \begin{tabular}{c|cc}
    \toprule
    Method & 20\% & 50\% \\
    \midrule
    0\% Baseline & \multicolumn{2}{c}{3.06}	 \\
    LLM-Pruner & 2.63\small{(-14\%)} & 2.17\small{(-29\%)} \\
    NutePrune & 2.72\small{(-11\%)} & 2.18\small{(-29\%)} \\
    \bottomrule
    \end{tabular}
    \caption{\label{table-speed}
Inference latency of pruned LLaMA-7B.
}
\end{table}

\paragraph{Training cost}
We report the memory and latency cost on different settings in Table \ref{table-cost}.
For extra GPU memory cost of PKD, NutePrune snapshot lightweight modules (masks and LoRA) of numerous teachers into CPU. Only one teacher module is loaded onto the GPU when needed, resulting in negligible memory cost compared with KD.
In terms of extra time cost, compared with  supervised training, KD requires one extra forward pass of teacher model, which is inevitable and cost 18.0\% extra latency. When snapshoting a teacher or switching to a new teacher, due to the extremely low frequency of operations, the time can be ignored. Introducing $\mathcal{L}_{layer}$ requires additional 32\% memory which is also efficient compared to conventional KD. 

\begin{table}[h!]
  \centering
  \begin{tabular}{ccc|cc}
    \toprule
    \small{Progressive} & KD & $\mathcal{L}_{layer}$ & Memory & Latency \\
    \midrule
    ~ & ~ & ~ & 27.68 & 3.67 \\
    ~ & $\checkmark$ & ~ & 28.67 & 4.33 \\
    $\checkmark$ & $\checkmark$ & ~ & 28.69 & 4.33 \\
    $\checkmark$ & $\checkmark$ & $\checkmark$ & 38.00 & 5.52 \\
    \bottomrule
    \end{tabular}
    \caption{\label{table-cost}
Training cost measured by average GPU memory (GB) and per step latency (s/iter).
}
\end{table}

\subsection{Ablation Study}
We validate the effectiveness of NutePrune and investigate which properties make for a good NutePrune. Results are average zero-shot performance with tuning but without post fine-tuning, unless otherwise stated.

\paragraph{Effectiveness of PKD}
To validate progressive knowledege distillation (PKD) in our NutePrune, we conduct ablation studies on various learning strategies. We eliminate the progressive schedule and adopt standard KD, where the intact model serves as the teacher throughout. Subsequently, we exclude the entire distillation procedure and employ the standard generative language model loss, specifically next-token prediction, to train masks and LoRA modules. The results presented in Table \ref{table-abl_prog} demonstrate the critical role of KD in enhancing performance, with further improvements achieved through PKD. This phenomenon is particularly pronounced at higher sparsity.

\begin{table}[h!]
  \centering
  \begin{tabular}{cc|cc}
    \toprule
    Progressive & KD & 20\% & 50\% \\
    \midrule
    $\checkmark$ & $\checkmark$ & \textbf{63.57} & \textbf{53.94}\\
    ~ & $\checkmark$ & 63.19 & 52.73\\
     ~ & ~ & 59.98 & 41.77\\
    \bottomrule
    \end{tabular}
    \caption{\label{table-abl_prog}
NutePrune and variants at 20\%/50\% sparsity.
}
\end{table}


\paragraph{Two stages of PKD}
PKD includes one stage before reaching target sparsity and the other stage after that. Different progressive schedules are adopted. To assess the effectiveness of them, we conducted an ablation study at 50\% sparsity under two training settings, as shown in Table \ref{table-abl_twostage}: training masks only and co-training masks with LoRA. In a stage without a progressive schedule, the intact model serves as the teacher. For the masks-only scenario, adopting either stage 1 or 2 alone yields significant improvements over KD. And for co-training, significant improvement is observed when both stages are adopted simultaneously.
\begin{table}[h!]
  \centering
  \begin{tabular}{cc|cc}
    \toprule
    \multirow{2}{*}{Stage 1} & \multirow{2}{*}{Stage 2} & \multicolumn{2}{c}{Avg.(\%)} \\
    ~ & ~ & masks-only & co-train \\
    \midrule
    $\checkmark$ & $\checkmark$ & \textbf{53.14} & \textbf{53.94} \\
    $\checkmark$ & ~ & 52.31 & 52.79 \\
    ~ & $\checkmark$ & 52.40 & 52.53 \\
    ~ & ~ & 51.83 & 52.73 \\
    \bottomrule
    \end{tabular}
    \caption{\label{table-abl_twostage}
Performance of two stages of PKD.
}
\end{table}

\paragraph{Sparsity gap and interval of teachers}
During stage 1, the sparsity gap between teacher and student model is an important hyperparameter. As shown in Table \ref{table-abl_gap}, a 10\% gap is deemed appropriate to prevent a gap that is too small, as it may result in insufficient guidance, or a gap that is too large, which would toughing distillation. And when taking snapshots of students as teachers, it is preferable to save as many teachers as possible to facilitate more comprehensive training. However, it comes with extra costs. As demonstrated in Table \ref{table-abl_interval}, selecting an interval of 1\% leads to significant improvement over the 10\% interval, and the associated extra storage is acceptable. 
\begin{table}[h!]
  \centering
  \begin{tabular}{c|ccc}
    \toprule
    Sparsity Gap & 5\% & 10\% & 20\% \\
    \midrule
    Avg.(\%) & 53.62 & \textbf{53.94} & 53.04 \\
    \bottomrule
    \end{tabular}
    \caption{\label{table-abl_gap}Performance of various sparsity gap.}
\end{table}

\begin{table}[h!]
\vspace{-0.5cm}
  \centering
  \begin{tabular}{c|c|c}
    \toprule
    Snapshot Interval & CPU Storage & Avg.(\%) \\
    \midrule
    1\% (ours) & 728MB & \textbf{53.94} \\
    10\% & 73MB & 53.27 \\
    \bottomrule
    \end{tabular}
    \caption{\label{table-abl_interval}Storage and performance of various intervals.}
\end{table}

\section{Conclusion}
In this work, we propose NutePrune as a novel efficient progressive structured pruning method for LLMs. Our well-designed techniques minimize the memory cost of KD, enabling NutePrune to utilize numerous teachers to mitigate the capacity gap between teacher and student and improve the quality of distillation. We show the effectiveness of NutePrune across various base models on diverse metrics. This work contributes to structured pruning techniques for LLMs, particularly in resource-constrained scenarios.

\section{Limitations}
Recent work \cite{ma2023llm,xia2023sheared} proves that using extensive data for post-training could substantially enhance the performance, but it comes with a substantial increase in computational costs. We target on pruning on resource-constraint scenarios and leave pruning with extensive data for future work.

\bibliography{acl_latex}

\appendix

\section{Detailed $L_0$ Regularization}
\label{sec:appendix_l0}

To get the learnable masks $\mathbf{z}$, $L_0$ regularization introduces a sampling strategy as augmentation during training. $\mathbf{z}$ is a real number in $[0, 1]$ and is obtained from:

\begin{equation}
\begin{aligned}
\mathbf{u} & \sim U(0,1) \\
\mathbf{s} & =\operatorname{sigmoid}\left(\frac{1}{\beta} \log \frac{\mathbf{u}}{\mathbf{1 - u}}+\log \alpha\right) \\
\tilde{\mathbf{s}} & =\mathbf{s} \times(r-l)+l \\
\mathbf{z} & =\min (1, \max (0, \tilde{\mathbf{s}})),
\end{aligned}
\end{equation}
where $\mathbf{u}$ is uniformly sampled between 0 to 1. $\alpha$ is the parameter to be learned and $\beta$ is a hyperparameter. $l, r$ is often $-0.1$ and $1.1$ to ensure most $\mathbf{z}$ are either $0$ or $1$ after training.

To prevent models from drastically converging to different sizes, we follow \cite{wang2019structured} to use this Lagrangian term:
\begin{equation}
    \mathcal{L}_0=\lambda_1 \cdot(\hat{s}-t)+\lambda_2 \cdot(\hat{s}-t)^2,
\end{equation}
where $\lambda_1$ and $\lambda_2$ are both learnable. This loss term $\mathcal{L}_0$ will impose $\hat{s}$ to gradually converge to target sparsity $t$.

\section{Zero-shot Performance with Newer Version}
\label{sec:appendix_new}

\textit{lm-evaluation-harness} released a new version in June 2023 to assess the zero-shot performance of LLaMA \footnote{https://github.com/EleutherAI/lm-evaluation-harness/pull/531}. This update addressed a tokenization bug specific to LLaMA, resulting in higher and more accurate performance results compared to the older version. Despite these improvements, current state-of-the-art reports continue to reference the older version. Consequently, we conducted experiments using both the new and old versions, and the detailed results for the new version are presented in Table \ref{table-main_new}.

\begin{table*}
  \centering
  \small
  \begin{threeparttable}
  \begin{tabular}{ccc|ccccccc|c}
    \toprule
    Ratio & Tune & Method & BoolQ & PIQA & HellaSwag & WinoGrande & ARC-e & ARC-c & OBQA & $\star$Avg. \\
    \midrule
    0\% & ~ & LLaMA-7B & 75.11 & 79.16 & 76.21 & 69.85 & 75.29 & 44.71 & 44.40 & 66.39 \\
        \hline
    \multirow{2}{*}{20\%} & ~  & LLM-Pruner & 57.49 & 76.06 & 69.53 & 63.93 & 67.17 & 38.05 & 40.80 & 59.01 \\
    ~ & ~ & $\dagger$\textbf{NutePrune} & 70.21 & 76.93 & 71.66 & 68.27 & 71.09 & 40.44 & 42.60 & \textbf{63.03} \\
    \hline
    \multirow{6}{*}{20\%} & \multirow{6}{*}{$\checkmark$}   
     & MaP & 64.43 & 76.82 &	67.42 &	63.61 &	66.92 &	36.95 &	44.20 &	60.05 \\
      ~ & ~ & MvP & 64.56	& 75.19	 & 64.71	& 64.09 & 	66.12 &	38.65 &	38.00	& 58.76 \\
      ~ & ~ & LLM-Pruner & 67.37 & 77.86 & 71.47 & 65.90 & 69.57 & 39.59 & 41.80 & 61.94 \\
    ~ & ~ & Compresso & 73.21 & 75.90 & 66.90 & 68.90 & 69.99 & 41.47 & 41.80 & 62.60 \\
    ~ & ~ & $\ddagger$\textbf{NutePrune} & 73.79 & 77.37 & 72.27 & 67.48 & 72.77 & 38.91 & 42.40 & 63.57 \\
    ~ & ~ & \textbf{NutePrune} & 75.38 & 78.02 & 72.97 & 67.40 & 73.82 & 40.36 & 42.80 & \textbf{64.39} \\
    \hline
    25\% & ~ & $\dagger$\textbf{NutePrune} & 71.53 &	76.50 & 70.60 & 65.98 & 69.11 & 39.93 & 38.40 & 61.72 \\
    \hline
    \multirow{2}{*}{25\%} & \multirow{2}{*}{$\checkmark$} & $\ddagger$\textbf{NutePrune} & 72.91 & 77.42 & 70.34 & 68.11 & 70.92 & 41.55 & 40.60 & 63.12 \\
    ~ & ~ & \textbf{NutePrune} & 74.95 & 77.75 & 71.27 & 67.40 & 71.25 & 41.81 & 42.00 & 63.78 \\
    \hline
    \multirow{2}{*}{50\%}& ~   & LLM-Pruner & 46.48 & 61.10 & 36.87 & 51.78 & 35.10 & 27.65 & 27.60 & 40.94 \\
    ~& ~  & $\dagger$\textbf{NutePrune} & 65.38 & 69.04 & 55.08 & 61.33 & 55.72 & 30.80 & 34.60 & \textbf{53.14} \\
    \hline
    \multirow{6}{*}{20\%} & \multirow{6}{*}{$\checkmark$}   
     & MaP & 43.33	& 67.46 &	44.27 &	54.78 &	52.19 &	30.46 &	31.80 &	46.33 \\
      ~ & ~ & MvP & 60.00	& 63.11 &	41.73 &	56.04 &	47.10 &	27.05 &	28.60 &	46.23 \\
      ~ & ~ & LLM-Pruner & 57.89 & 69.97 & 50.06 & 52.64 & 49.66 & 28.58 & 34.00 & 48.97 \\
    ~ & ~ & Compresso & 61.31 & 66.32 & 40.73 & 52.41 & 51.18 & 27.65 & 32.40 & 47.43 \\
    ~ & ~ & $\ddagger$\textbf{NutePrune} & 67.25 & 70.67 & 56.64 & 59.83 & 57.07 & 31.74 & 34.40 & 53.94  \\
    ~ & ~ & \textbf{NutePrune} & 67.52 & 71.60 & 58.64 & 60.14 & 59.72 & 32.94 & 33.80 & \textbf{54.91} \\
    \bottomrule
    \end{tabular}
    \end{threeparttable}
    \caption{\label{table-main_new}Zero-shot performance of the compressed LLaMA models in the new version of lm-evaluation-harness. \textbf{Bold} denotes the best average performance at the same setting.}
\end{table*}

\section{Pruning at Higher Sparsity}
To demonstrate the effectiveness of NutePrune at higher sparsity, we conducted experiments at 70\% sparsity in Table \ref{table-main_new}.

\begin{table*}[h!]
  \centering
  \small
  \begin{threeparttable}
  \begin{tabular}{cc|c|ccp{0.4in}p{0.5in}p{0.35in}p{0.35in}p{0.33in}|cc}
    \toprule
    Ratio  & Method & WikiText2$\downarrow$ & BoolQ & PIQA & HellaSwag & WinoGrande & ARC-e & ARC-c & OBQA & $\star$Avg.\\
    \midrule
    0\% & LLaMA-7B & 5.68 & 75.11 & 79.16 & 76.21 & 69.85 & 75.29 & 44.71 & 44.40 & 66.39 \\
        \hline
    \multirow{2}{*}{70\%} &  LLM-Pruner & 56.33 & 47.28 & 60.83 & 31.66 & 50.75 & 39.56 & 24.83 & 28.80 & 40.53  \\
    ~  & \textbf{NutePrune} & \textbf{34.30} & 62.08 & 62.30	 & 39.43 & 51.46 & 42.17 & 26.19 & 30.20	& \textbf{44.83}
 \\
    \bottomrule
    \end{tabular}
    \end{threeparttable}
    \caption{\label{table-70}Performance (\%) of the compressed LLaMA-7B models at 70\% sparsity.}
\end{table*}

\section{Pruned Structure}
To gain insights into the pruned model, we present a detailed overview of the pruned structure at sparsity levels of 20\% and 50\%. The original hidden dimension is 4096, with a number of heads set at 32 and an intermediate dimension of 11008. Tables \ref{table-structure_20} and \ref{table-structure_50} reveal several observations. Notably, NutePrune tends to avoid pruning the hidden dimension, which aligns with the observation that pruning it may result in significant performance degradation \cite{ma2023llm}. Regarding heads and intermediate dimensions, NutePrune tends to prune the the last few layers. This observation differs from LLM-Pruner, which asserts the importance of the last layers. Further analysis of this phenomenon is left for future work.

\begin{table*}
  \centering
  \begin{threeparttable}
  \begin{tabular}{ccccccccc}
    \toprule
    \# Hidden Dim & \multicolumn{8}{c}{4070} \\
    \hline
    Layer & 1 & 2 & 3 & 4 & 5 & 6 & 7 & 8 \\
    \# Head & 23 & 22 & 30 & 22 & 29 & 27 & 30 & 28 \\
    \# Intermediate Dim  & 5832 & 7820 & 9169 & 9187 & 8967 & 9163 & 9186 & 9112  \\
    \hline
    Layer & 9 & 10 & 11 & 12 & 13 & 14 & 15 & 16 \\
    \# Head  & 30 & 30 & 31 & 30 & 32 & 27 & 30 & 30 \\
    \# Intermediate Dim  & 9261 & 9165 & 9303 & 9695 & 10005 & 10258 & 10417 & 10564  \\
    \hline
    Layer & 17 & 18 & 19 & 20 & 21 & 22 & 23 & 24 \\
    \# Head  & 30 & 29 & 26 & 25 & 23 & 21 & 16 & 21 \\
    \# Intermediate Dim  & 10715 & 10759 & 10785 & 10790 & 10808 & 10778 & 10729 & 10707  \\
    \hline
    Layer & 25 & 26 & 27 & 28 & 29 & 30 & 31 & 32 \\
    \# Head  & 14 & 15 & 6 & 7 & 11 & 8 & 7 & 9   \\
    \# Intermediate Dim  & 10568 & 10366 & 10054 & 9403 & 8519 & 7297 & 6588 & 5164 \\
    \bottomrule
    \end{tabular}
    \end{threeparttable}
    \caption{\label{table-structure_20}Detailed structure of compressed 20\% LLaMA.}
\end{table*}

\begin{table*}
  \centering
  \begin{threeparttable}
  \begin{tabular}{ccccccccc}
    \toprule
    \# Hidden Dim & \multicolumn{8}{c}{4021} \\
    \hline
    Layer & 1 & 2 & 3 & 4 & 5 & 6 & 7 & 8 \\
    \# Head & 24 & 21 & 27 & 18 & 26 & 21 & 20 & 20 \\
    \# Intermediate Dim  & 3980 & 6216 & 7120 & 6590 & 5889 & 5731 & 5283 & 4944  \\
    \hline
    Layer & 9 & 10 & 11 & 12 & 13 & 14 & 15 & 16 \\
    \# Head  & 25 & 21 & 27 & 26 & 23 & 21 & 25 & 21  \\
    \# Intermediate Dim  & 4879 & 4563 & 4590 & 5040 & 5469 & 5832 & 6215 & 778   \\
    \hline
    Layer & 17 & 18 & 19 & 20 & 21 & 22 & 23 & 24 \\
    \# Head  & 23 & 23 & 21 & 19 & 16 & 15 & 6 & 13  \\
    \# Intermediate Dim  & 7706 &8121 &8005 &8091 &8232 &7705& 6879 &6134   \\
    \hline
    Layer & 25 & 26 & 27 & 28 & 29 & 30 & 31 & 32 \\
    \# Head  & 7 & 8 & 2 & 5 & 7 & 5 & 3 & 8   \\
    \# Intermediate Dim  & 4912 &3762 &3107 &2527 &2357 &2314 &2931 &2622 \\
    \bottomrule
    \end{tabular}
    \end{threeparttable}
    \caption{\label{table-structure_50}Detailed structure of compressed 50\% LLaMA.}
\end{table*}

\section{Generated Examples}

We present generated examples from our pruned model using NutePrune at 20\% sparsity. We provide examples of three types: without tuning (w/o tune), with tuning but without post-finetuning (w/ tune), and with tuning and post fine-tuning (w/ tune + post FT). The results are displayed in Table \ref{table-examples}.


\begin{table*}[h!]
    \centering
    \resizebox{\linewidth}{!}{
    \tiny
    \begin{tabular}{p{0.8cm}|p{9.8cm}}
        \toprule
        Model & Generated Sentences \\
        \midrule
        \multirow{16}{*}{w/o Tune} & \underline{Neural Sequential Model, especially transformers,} are used to model the human brain. The model is based on the assumption that the brain is a system of interconnected modules, each of which is responsible for a specific function. The modules are connected by synapses, which are the synaptic connections between neurons. The synapsis are responsible for the transfer of information between the modules. The information is transferred in the form of a signal, which is a combination of the input signal and the output signal. The input signal is the signal that is received by the module, while the output is the output of the module. The output signal is then passed on to the next ... \\
        \cmidrule{2-2}
        & \underline{AI can create a logo in seconds} . It can also create a logo that is more creative than a human designer. AI can also create logos that are more creative and unique than a human can. AI is also able to create logos in a short time. AI has the ability to create logotypes. AI will be able to create a logo for a company in a short period of time. A logo is a symbol that is used to identify a company. A logo can be used to identify the company. AI also has the ability of creating logos that can be used in different ways. AI creates logos that have ... \\
        \cmidrule{2-2}
        & \underline{What’s great about the holiday season} ? The holiday spirit, of course! But it’s also a time for giving. And giving is what we do best at the Chamber. We give you the opportunity to give back to your community. We’ve got a great list of charitable organizations that are looking for your help. And we’ve made it easy for you to give. Just click on the link below and you’ll be taken to the website of the charity of your choice. You can make a donation online, or you can mail a check. Either way, you’re giving back to the community. And that’s what the ... \\
        \midrule
        \multirow{16}{*}{\parbox{0.8cm}{w/ Tune}} & \underline{Neural Sequential Model, especially transformers,} are used to model the human brain. The model is a neural network that is trained to recognize patterns in data. The network is trained by feeding it data and then testing it on new data. This is a form of machine learning. The neural network is a type of artificial neural network. The term neural network refers to a group of neurons that are connected together. The neurons are connected by synapses. The synapsis are the connections between the neurons. The connections are called weights. The weights are used by the network to determine how much influence each neuron has on the other neurons in the network ... \\
        \cmidrule{2-2}
         & \underline{AI can create a logo in seconds} , but it can’t design a logo that will work for a company. AI is great for creating a logo, but not for designing a logo. A logo is a visual representation of a company’s brand. It’s a visual symbol that represents the company’ logo. It is a symbol that is used to identify a company and its products. A good logo is one that is memorable and easy to recognize. A great logo is memorably and easy recognizable. A bad logo is not memorable or easy to recognize. A poorly designed logo is also not memorably or easy recognized. A ...\\
        \cmidrule{2-2}
         & \underline{What’s great about the holiday season} is that it’s a time for giving. And what better way to give than to give the gift of a new home? Whether you’re buying a home for the first time or you’ve been in the market for a while, there are a few things you should know before you start looking. 1. Know your budget. This is the most important thing to consider when you’ll be looking for a home. You’ll want to know how much you can afford to spend on a home, and you’d want to make sure you don’t go over your budget when you start searching for homes ... \\
        \midrule
        \multirow{17}{*}{\parbox{0.8cm}{w/ Tune + Post FT}} & \underline{Neural Sequential Model, especially transformers,} is a popular model for deep learning. It is a model that can be trained on large datasets. It is used for tasks such as image classification, natural language processing, and other tasks. It is an effective model for tasks such  as image classification and natural language processing. It is also used for tasks like natural language understanding and natural ﬁlm ﬁction. It is able to learn from large datasets and can be trained for various tasks. It can be used for tasks that require a large amount of data. It is capable of learning from large datasets.  It is used to train neural networks for tasks such image classification ... \\
        \cmidrule{2-2}
         & \underline{AI can create a logo in seconds}. AI Logo Designer. A Logo AI Can Create. A logo is a key element of any business. It helps to create a unique identity and helps to stand out from the crowd. A good logo can help to build trust and credibility. It can also help to increase the sales. Alogoai.com is a AI logo design service. We can design logos for any business, with the latest technology. We provide high quality logos at affordable prices. Logo design service provided by Alogo AI. Logos designed by AI are more accurate and effective. Logs designed by ...\\
        \cmidrule{2-2}
         & \underline{What’s great about the holiday season} is the opportunity to give back to the community. Whether it’s donating to a local charity or giving to a needy family, it’ll make you feel good to know that you’re making a difference. Here are some ideas to get started. Volunteer at a local organization. If you have some free time, consider volunteering at a nonprofit organization. You can help with the holidays by helping with the decorations, helping with food preparation, or even helping with some of the administrative tasks. You’ll feel great knowing that you are making a contribution to the organization. Donate ... \\
        \bottomrule
    \end{tabular}
    }
    \caption{Generated Examples from the Compressed LLaMA-7B at 20\% sparsity} \label{table-examples}
\end{table*}

\end{document}